\documentclass[10pt,twocolumn,letterpaper]{article}

\usepackage[pagenumbers]{cvpr} % To force page numbers, e.g. for an arXiv version

\newcommand{\ignore}[1]{}

\definecolor{cvprblue}{rgb}{0.21,0.49,0.74}
\usepackage[pagebackref,breaklinks,colorlinks,allcolors=cvprblue]{hyperref}

\makeatletter
\def\blfootnote{\gdef\@thefnmark{}\@footnotetext}
\makeatother

\def\paperID{***} % *** Enter the Paper ID here
\def\confName{CVPR}
\def\confYear{2026}

\title{Goal Force: Teaching Video Models To Accomplish Physics-Conditioned Goals}

\author{Nate Gillman$^{1}$ \hspace{.1cm} Yinghua Zhou$^{1}$ \hspace{.1cm} Zitian Tang$^{1}$ \hspace{.1cm} Evan Luo$^{1}$ \hspace{.1cm} Arjan Chakravarthy$^{1}$ \hspace{.1cm}\\ \hspace{.1cm} Daksh Aggarwal$^{1}$ \hspace{.1cm} Michael Freeman$^{2}$ \hspace{.1cm} Charles Herrmann$^{1}$ \hspace{.1cm} Chen Sun$^{1}$ \\[1ex]
Brown University$^{1}$ \quad Cornell University$^{2}$\\[.7ex]
{\small\nolinkurl{{nate_gillman, yinghua_zhou, zitian_tang, chensun}@brown.edu}}}
\begin{document}
\maketitle
% \blfootnote{Correspondence to: {\tt \{nate\_gillman,chensun\}@brown.edu}.}
% \blfootnote{Correspondence to: \nolinkurl{{nate_gillman, chensun}@brown.edu}.}

\begin{abstract}
Recent advancements in video generation have enabled the development of ``world models'' capable of simulating potential futures for robotics and planning. However, specifying precise goals for these models remains a challenge; text instructions are often too abstract to capture physical nuances, while target images are frequently infeasible to specify for dynamic tasks. To address this, we introduce Goal Force, a novel framework that allows users to define goals via explicit force vectors and intermediate dynamics, mirroring how humans conceptualize physical tasks. We train a video generation model on a curated dataset of synthetic causal primitives—such as elastic collisions and falling dominos—teaching it to propagate forces through time and space. Despite being trained on simple physics data, our model exhibits remarkable zero-shot generalization to complex, real-world scenarios, including tool manipulation and multi-object causal chains. Our results suggest that by grounding video generation in fundamental physical interactions, models can emerge as implicit neural physics simulators, enabling precise, physics-aware planning without reliance on external engines.
We release all datasets, code, model weights, and interactive video demos at our project page, \href{https://goal-force.github.io/}{https://goal-force.github.io/}.
\end{abstract}

\begin{figure}[t]
  \centering
   \includegraphics[width=\linewidth]{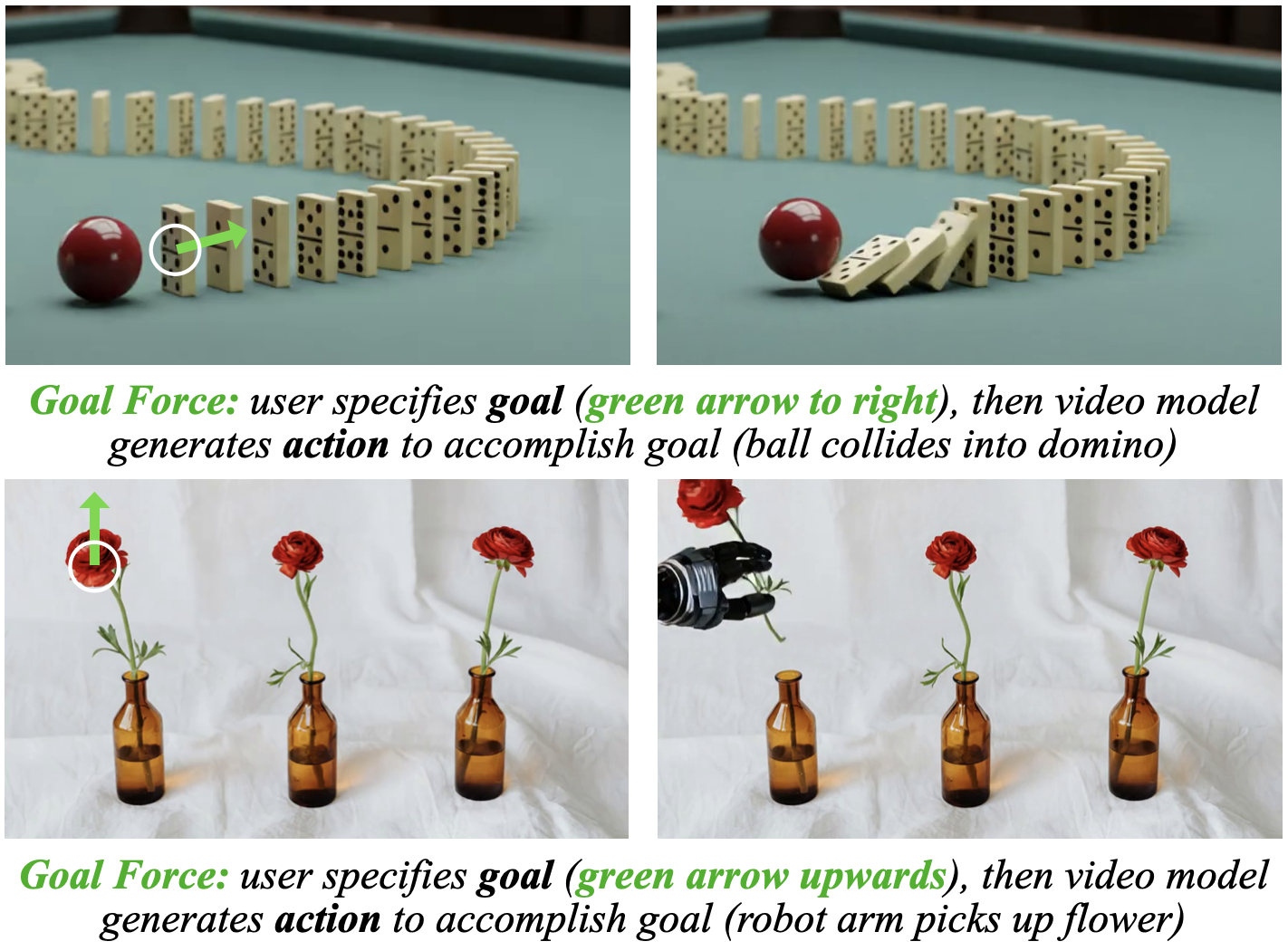}
   \caption{Given a force-conditioned task, \textbf{goal force} enables video models to generate the antecedent action to accomplish the task.}
   \label{fig:mini_teaser}
\end{figure}

\begin{figure*}[t]
  \centering
   \includegraphics[width=\linewidth]{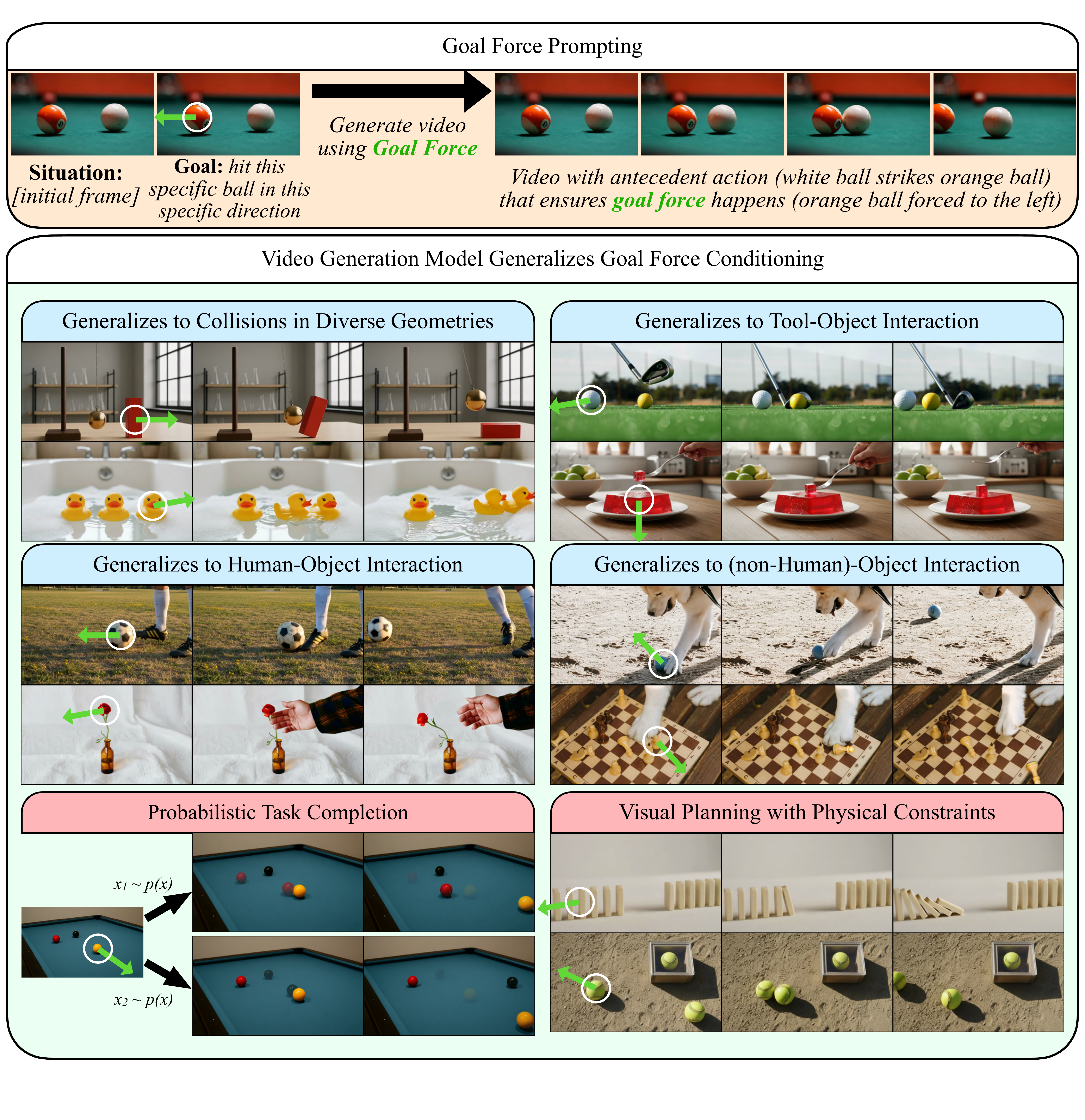}
   \vspace{-4em}
   \caption{
   \textbf{Goal Force}: A user provides an input image and a \textbf{\textcolor[RGB]{31,166,3}{goal force}}, and the model generates a video containing a force that locally causes the goal force. 
   Our model generalizes to diverse objects and interactions and enables visual planning, respecting the physical properties of the objects and their environments.\vspace{-1.5em}}
   \label{fig:teaser}
\end{figure*}

\section{Introduction}
\label{sec:intro}

The past two years have witnessed a paradigm shift in video generation, evolving from coarse, rudimentary clips to near-photorealistic sequences~\cite{sora2024, bar2024lumiere, agarwal2025cosmos}. This progress has sparked considerable interest in leveraging these models as ``world models" for robotics and planning. One of the most exciting possibilities for using ``world models'' in planning involves generating a video that transitions from a current state (an initial frame) towards a specified goal state~\cite{Guo2025-oy, ko2024avdc}. Consider a soccer player at the start of a game: the initial frame shows the ball at midfield, and the objective is to score. Existing approaches predominantly rely on text or static images to define these goals. However, for complex physical tasks involving multi-step dynamics, these modalities often prove insufficient. Text is frequently too abstract; a soccer player's intent is rarely just to ``shoot at the goal," but rather to strike the ball with specific force and precision. Conversely, specifying a goal via a target image is often overly burdensome or infeasible, potentially requiring a user to render the exact lighting of a ball entering the net.

In contrast, humans approach tasks differently than through abstract text or pixel-perfect images alone. We often decompose long, abstract tasks into concrete sub-goals that, particularly in sports, possess distinct physical properties like spatial location, dynamics, and motion. When taking a penalty kick, a soccer player does not focus merely on the static end state of the ball in the net, nor do they simply rely on the abstract concept of scoring. Instead, they aim to impart a specific trajectory and velocity—a ``goal force''—onto the ball. This paper proposes a method that aligns with this intuition: defining goals through desired forces and intermediate dynamics. By specifying these goal forces, rather than limiting users to static endpoints or requiring direct, low-level scene manipulation, we offer a mechanism that is both precise enough for physics-based planning and intuitive for human users.

To accomplish this, we introduce a framework that conditions video generation on explicit goal force vectors. We curate a dataset of paired videos and ``goal forces,'' adapting a state-of-the-art open-source video model to accept these forces as a  control signal. Our training strategy relies on the hypothesis that learning fundamental physical interactions can bootstrap complex reasoning. We train the model on simple, synthetic examples of causal primitives, such as elastic collisions and falling dominos. Crucially, we find that this grounded training enables \textit{non-trivial generalization} to highly diverse scenarios (Figure~\ref{fig:teaser}).

Our empirical results demonstrate that the model learns to propagate forces through time and space, handling chains of events where one object exerts force on another, which in turn influences a third. Remarkably, this capability extends to zero-shot tool usage; for instance, the model can infer how to use a golf club to impart the desired force onto a ball, and to pick up a rose via its stem as opposed to its petals (Figure~\ref{fig:mini_teaser}), despite only being trained on simpler collision data. This suggests the model is not merely memorizing patterns but acting as an implicit neural physics simulator.

Our main contributions are as follows:

\begin{enumerate}
\item We propose \emph{Goal Force}, a new task and model which teaches video models to plan a causal chain of physical interactions to achieve a specified goal force. This moves beyond prior direct-force methods and changes how goals can be specified in world models.

\item We propose a training paradigm with a novel multi-channel control signal (for goal forces, direct forces, and mass) that teaches the model to act as an implicit neural physics simulator, requiring no simulator at inference.

\item We demonstrate powerful out-of-domain generalization: despite training only on simple synthetic data (\eg, balls, dominos), our model leverages the base video model's rich  prior to generate complex, physically-plausible scenarios involving tool use, human-object interaction, and intricate multi-object collisions.

\end{enumerate}

We release training and evaluation code, model weights, synthetic training data, and benchmark datasets at our project page, \href{https://goal-force.github.io/}{https://goal-force.github.io/}.

\begin{figure*}[t]
  \centering
   \includegraphics[width=.95\linewidth]{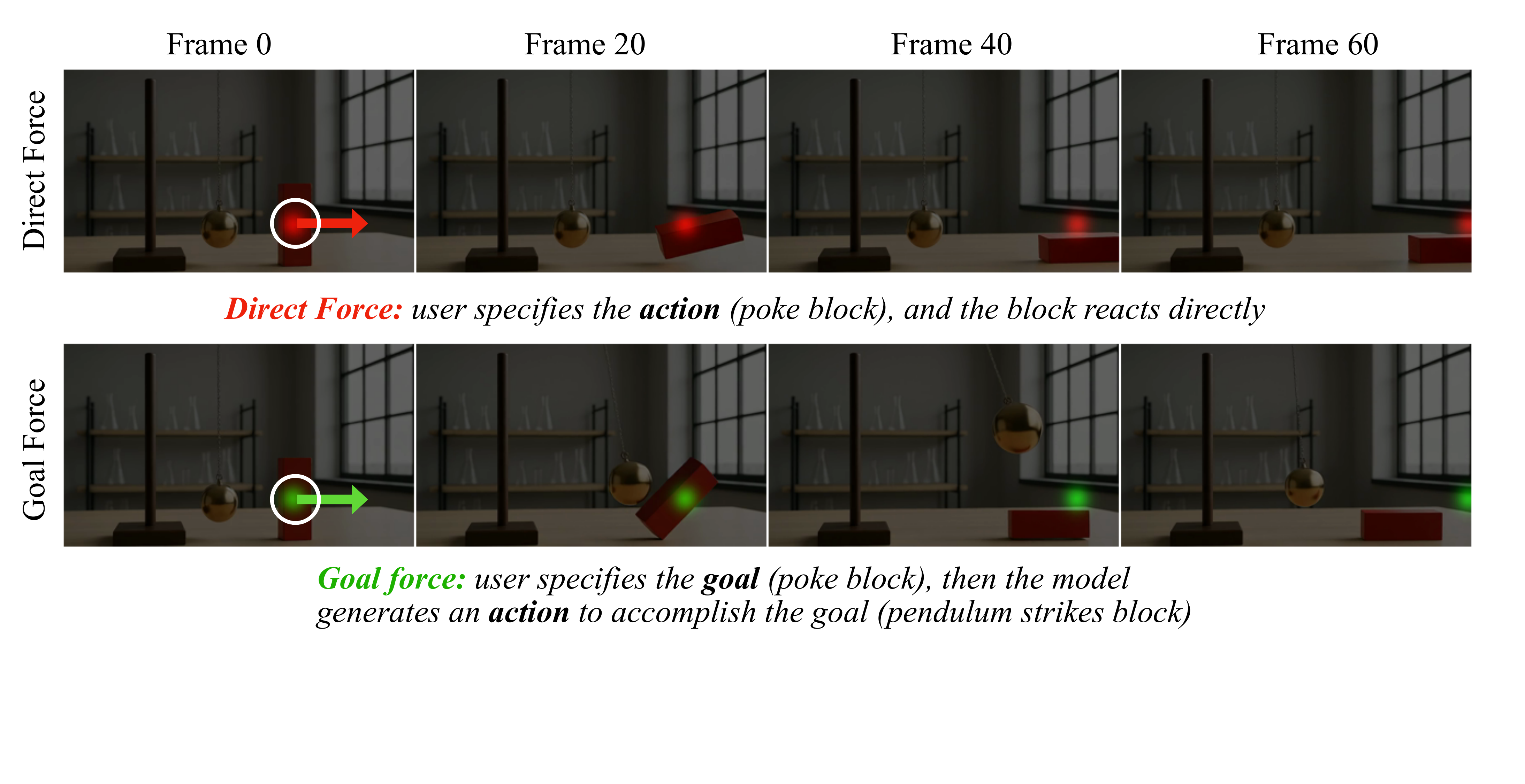}
   \vspace{-50pt}
   \caption{\textbf{Force prompt and goal force prompt result in different behaviors.}
    With a \textbf{\textcolor[RGB]{229,0,13}{direct force}} applied to the red block (top), the effect is directly  materialized (i.e. the block falls over). 
    The force in this case is encoded in the red channel of the control signal as a moving Gaussian blob.
    In contrast, with a \textbf{\textcolor[RGB]{31,166,3}{goal force}} applied to the red block (bottom), the model must find the antecedent motion to achieve the goal force (i.e. the pendulum swings to knock over the block). 
    The force in this case is encoded in the green channel of the control signal as a moving Gaussian blob.
    We visualize the control signal overlaid on top of the video via alpha blending.\vspace{-1em}}
   \label{fig:control_signal}
\end{figure*}

\section{Related Works}
\label{sec:related_works}

\textbf{Video generative models:} In recent years, video generation models have achieved remarkable progress in visual fidelity and the plausible rendering of complex dynamics \citep{makeavideo2022, imagenvideo2022, svdblattmann2023, emuvideo2023, bar2024lumiere}. 
The introduction of models like Sora \citep{sora2024} highlighted the potential for using large-scale generative models as ``world simulators'' capable of rendering diverse physical phenomena. 
This progress has been mirrored in open-source efforts \citep{yang2024cogvideox, wan2025, Yang2025-iy}, which are increasingly approaching the quality of closed-source systems. 
While these models serve as powerful video priors, they are typically conditioned on text or images and lack interfaces for fine-grained, precise control over physical actions or interactions, which is a gap our work aims to address.

\noindent\textbf{Controllable video generation:} To address the need for greater control, many methods have been proposed. 
A significant portion of this research focuses on controlling the camera perspective \citep{he2024cameractrl, zheng2024cami2v, sun2024dimensionx}. 
Another major direction is motion control, which uses various paradigms like drag-based editing \citep{yin2023dragnuwa, wu2024draganything}, trajectory specification \citep{chen2023mcdiff, zhang2025tora, namekata2024sg}, or optical flow guidance \citep{shi2024motion, niu2024mofa, li2024image}. 
A limitation of many of these techniques \citep{yin2023dragnuwa, zhang2025tora} is their reliance on densely specified, complete trajectories, which makes them unsuitable for predictive tasks where the full motion is unknown. 
Prior work like Motion Prompting \citep{geng2024motion} allows for sparse trajectory inputs, but this still specifies motion rather than its underlying cause. 
More recently, Force Prompting \citep{gillman2025forcepromptingvideogeneration} introduced direct physical control by specifying a force vector. 
However, all these methods focus on direct, immediate interventions. 
Our work, \emph{Goal Force}, moves beyond this by enabling the model to reason about and plan a causal chain of forces: for example, hitting ball A in order to achieve a desired goal force on ball B.

\noindent\textbf{Physics simulators and hybrid approaches:} There is a long history of attempting to model physics from video. 
Early work \citep{doi:10.1073/pnas.1306572110,li2024generative} focused on extracting intuitive physical properties, such as the modal bases of vibrating objects, but these methods struggle to represent general motion. 
An alternative research line incorporates explicit physics simulation \citep{chen2022virtual, zhong2024reconstruction, le2023differentiable, xie2024physgaussian, zhang2024physdreamer, huang2024dreamphysics, liu2024physics3d, lin2024phys4dgen, aira2024motioncraft, physctrl2025,NIPS2015_d09bf415}. 
While physically accurate, these approaches generally require access to 3D geometry, which is often unavailable.
Hybrid models represent a compromise, as they combine physics simulators for dynamics with generative models for appearance \citep{liu2024physgen, tan2024physmotion, li2025wonderplay}.
A key limitation is that these models are constrained by the capabilities of their internal simulator (\eg, rigid bodies only) and require it at inference time.
More recent works have removed this dependency on internal simulators \citep{Watanabe2025-sw,Romero2025-kt} and can learn better representations of physical properties \citep{Ji2025-ou, Zhan2025-xo}, but these works focus on local physical properties rather than causal interactions.
Concurrent works have also explored using simulated data to fine-tune models for freefall \citep{li2025pisa} or learning 3D trajectories \citep{physctrl2025}. 
Our approach differs fundamentally: we do not use any physics simulator at inference time. Instead, we train the generative model itself to act as an approximate ``neural simulator'' that can reason about and plan causal interactions to achieve a specified goal.

\noindent\textbf{Interactive world models:} The concept of a ``world model'' \citep{ha2018world, NIPS2017_4c56ff4c} that can learn to simulate and interact with an environment has gained significant traction. 
To date, investigations have largely concentrated on video game environments \citep{valevski2024diffusion, che2024gamegen, bruce2024genie, Kang2025-ue}. 
While some recent studies have begun to explore real-world applications \citep{bar2024navigation, agarwal2025cosmos, Li2025-st, Zhang2025-eb, Guo2025-oy}, the forms of interaction are typically limited to text prompts or camera navigation. 
In contrast, our work introduces a new, physically-grounded form of interaction. 
By allowing a user to specify a goal force, we push the model to reason about physical cause-and-effect and plan the antecedent actions (like tool use or multi-object collisions) necessary to achieve that goal, representing a step towards more capable and physically-aware interactive world models.

\noindent\textbf{Planning with videos:} Video models have been applied to solve decision-making problems in robotic applications~\cite{ McCarthy2024TowardsGR, liang2024dreamitate}. A video generative model can serve as reward functions~\cite{ alejandro2023viper, huang2023diffusion}, dynamics models~\cite{yang2023learning, valevski2024diffusion}, and pixel-based planners~\cite{ko2024avdc, ajaj2023hip, Zhou2024RoboDreamerLC}. For example, UniPi~\cite{du2024video} and Adapt2Act~\cite{luo2025solving} employ text-conditioned video generative models to predict visual plans that depict future outcomes, which are then converted into robotic actions with inverse dynamics models. With our introduced framework, such visual planners can take goal forces, in addition to text, to specify the desired goals.

\section{Method: Prompting with Goal Force}
\label{sec:method}

Our method reframes force-conditioned video generation from specifying a \emph{direct force} (\eg, ~\citep{gillman2025forcepromptingvideogeneration,zhang2024physdreamer,liu2024physgen}) to declaring a desired \emph{goal force}. 
Given a starting frame $\phi$ and a text prompt $\tau$, the user specifies a ``goal force'' on a target object (make ball B move right). 
The model's task is to generate a video $v$ that synthesizes a physically-plausible \emph{antecedent causal chain} (ball A striking ball B) to achieve that goal.

We achieve this by training a video generative model to act as an implicit neural physics planner. The core of our approach is a novel training paradigm built on a multi-channel physics control signal and a curriculum of synthetic data.

\begin{figure*}[t]
  \centering
   \includegraphics[width=\linewidth]{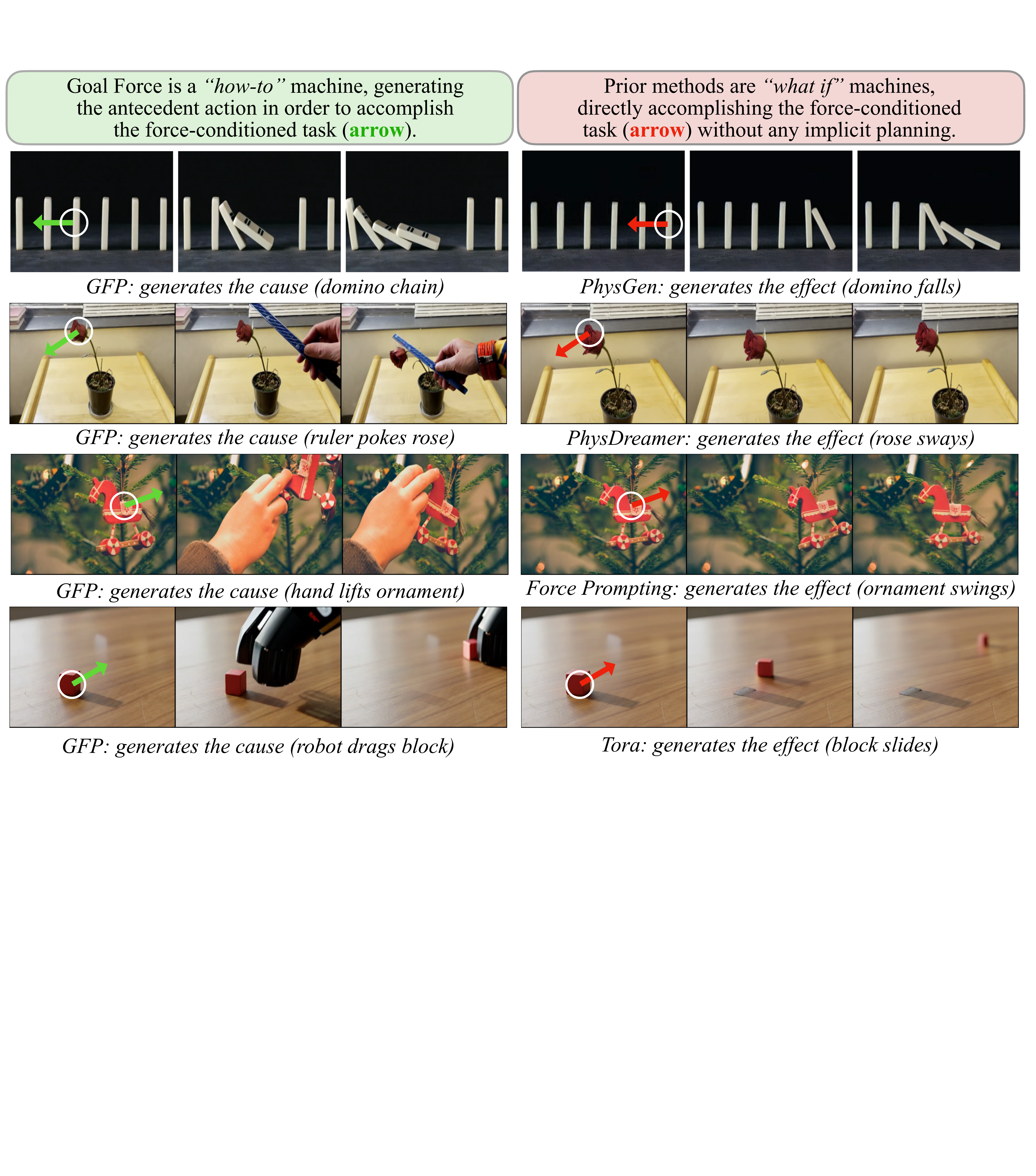}
    \vspace{-210pt}
   \caption{In prior methods (right), the user provides a force, and the model directly applies the force to the target object. 
   In our method (left), the user provides a goal force, and the model generates the causes that achieve the desired effect on the target object.
   The top three methods (PhysGen \citep{liu2024physgen}, PhysDreamer \citep{zhang2024physdreamer}, and Force Prompting \citep{gillman2025forcepromptingvideogeneration}) all accept forces as conditioning; the fourth method, Tora \citep{zhang2025tora}, accepts trajectories rather than forces, so we condition on an acceptable trajectory.\vspace{-1em}}
   \label{fig:prior_methods}
\end{figure*}

\subsection{Multi-Channel Physics Control Signal}
\label{subsec:control_signal}
We introduce a 3-channel physics control tensor $\tilde{\pi} \in \mathbb{R}^{f \times 3 \times h \times w}$, where $f$ is the number of frames, $h$ and $w$ are the spatial dimensions, and each of the 3 channels encodes a specific physical property. This tensor $\tilde{\pi}$ is the spatial-temporal encoding of the abstract user prompt.

\noindent\textbf{Channel 0: Direct Force.} Encodes an immediate, direct force (the ``cause''). Following \cite{gillman2025forcepromptingvideogeneration}, we represent this as a ``moving Gaussian blob'' video, where the blob's trajectory and duration are affinely proportional to the force vector (location, angle, and magnitude).

\noindent\textbf{Channel 1: Goal Force.} Encodes the desired \emph{outcome} (the ``effect'') on a target object. This channel uses the \emph{same} moving Gaussian blob representation to specify the desired force (and resulting motion) on the \emph{target} object.
We visualize the practical difference between a Goal force and a Direct force in Figure~\ref{fig:control_signal}.

\noindent\textbf{Channel 2: Mass.} Encodes privileged physical information, such as relative object mass. We represent this as a \emph{static} Gaussian blob in this channel, centered on the object, with a radius affinely proportional to its mass. The mass signal is optional, and offers an interface for users to provide more fine-grained, object-level physical properties, when they are available. When not provided, Goal Force can instead resort to the physical priors encoded in video generative models themselves, a behavior referred to as ``mass understanding'' in~\cite{gillman2025forcepromptingvideogeneration}.

\noindent\textbf{Force and Mass Normalization.}
We note that force and mass values are not calibrated to an absolute physical scale. Instead, they follow an intuitive, relative scale normalized \emph{within} each synthetic dataset (dominos, balls, plants). 
Our model learns this relative concept, as the Gaussian blob encoding is also defined proportionally to the value range of a given domain. This allows the model to generalize the \emph{idea} of force (\eg, ``small poke'' vs. ``large poke'') without requiring a unified, absolute scale.

\subsection{Goal Reaching via Implicit Planning}
\label{subsec:training_paradigm}
We train the model on a synthetic dataset of simple causal chains (colliding balls, falling dominos) and complex dynamics (swaying flowers), generated using Blender and PhysDreamer \citep{zhang2024physdreamer}. 
This dataset contains three scenarios:
\begin{itemize}
    \item \textbf{Dominos (3k videos):} Generated in Blender, these videos show a line of dominos where a direct force on one initiates a chain reaction, linking the ``cause'' to a ``goal force'' on a downstream domino.
    \item \textbf{Rolling Balls (6k videos):} Blender scenes of multiple balls. A direct force is applied to a ``projectile'' ball, which is aimed to either collide with a ``target'' ball (4.5k videos) or miss it (1.5k videos).
    \item \textbf{PhysDreamer Carnation (3k videos):} Videos of a flower swaying after being poked, generated with PhysDreamer \citep{zhang2024physdreamer}, a method that integrates 3D Gaussians and a physics simulator. This component teaches the model complex, non-rigid dynamics from a direct force.
\end{itemize}
Full data generation details are in Appendix~\ref{subsec:synthetic-extended}.

\begin{table*}
\begin{center}
\small % Smaller font size than \small
\setlength{\tabcolsep}{3pt} % Reduce column spacing (default is 6pt)
\begin{tabular}{l || ccc || ccc || ccc || ccc} 
Benchmark & 
\multicolumn{3}{c}{Two Object Collision} & 
\multicolumn{3}{c}{Multi Object Collision} & 
\multicolumn{3}{c}{Human Object Interaction} &  
\multicolumn{3}{c}{Tool Object Interaction} \\ \hline\hline
& \emph{Force} & \emph{Real.} & \emph{Visual} 
& \emph{Force} & \emph{Real.} & \emph{Visual}
& \emph{Force} & \emph{Real.} & \emph{Visual}
& \emph{Force} & \emph{Real.} & \emph{Visual}\\ 
& \emph{Adh.} & \emph{Motion} & \emph{Qual.}
& \emph{Adh.} & \emph{Motion} & \emph{Qual.}
& \emph{Adh.} & \emph{Motion} & \emph{Qual.}
& \emph{Adh.} & \emph{Motion} & \emph{Qual.}\\ 
\hline\hline
Text-only, zero-shot & 73.4\% & 67.2\% & 66.0\% & 72.0\% & 69.0\% & 66.8\% & 70.5\% & 47.5\% & 48.9\% & 74.5\% & 61.6\% & 58.7\%\\ 
Text-only, fine-tuned & 66.9\% & 60.3\% & 60.7\% & 67.0\% & 57.5\% & 61.0\% & 56.8\% & 48.4\% & 50.4\% & 60.3\% & 60.3\% & 55.8\%\\
\hline\hline
\end{tabular}
\caption{\textbf{Human study comparing Goal Force method to text-only baselines.} 
Numbers indicate the percentage of human pairwise preferences for Goal Force Prompting over each text-only baseline on each benchmark dataset. 
The proposed model consistently yields superior goal force adherence against both baselines, with minimal degradation of motion realism and visual quality.\vspace{-1.8em}}
\label{tab:human_study}
\end{center}
\end{table*}

This synthetic data for the ball collisions and domino collisions provides ground-truth pairs of (direct force, resulting goal force). 
Our key training strategy is to \textbf{randomly mask the causal information}. 
For each training video, we provide \emph{either} the direct force (in Ch 0) \emph{or} the goal force (in Ch 1), zeroing out the other. 
(And in scenes without collisions, namely 1/4 of the ball scenes and all the plant scenes, we only provide the direct force in Ch 0.)
This forces the model to learn the physical reasoning:
\begin{itemize}
    \item \textbf{Goal $\rightarrow$ Plan:} Given a goal force, the model must infer and generate the antecedent direct-force event.
    \item \textbf{Action $\rightarrow$ Outcome:} Given a direct force, the model must simulate the resulting collision and secondary force.
\end{itemize}
The mass channel (Ch 2) is also randomly masked during training. This teaches the model to leverage privileged physics information when available but also to rely on its internal, learned physics prior to estimate properties (like mass) from appearance when it is not.
The text prompt's role is to set the semantic context (\eg, ``a pool table'') and guide the model toward a plausible distribution of videos. 
It does not, however, specify the low-level causal plan, such as which ball should strike another. This ambiguity is intentional: it forces the model to leverage its internal prior to plan a valid antecedent action, constrained only by the specific objective of the goal force prompt.

\subsection{Architecture and Training Details}
\label{subsec:architecture}
We build our model on Wan2.2 \citep{wan2025}, a Mixture-of-Experts diffusion model. 
We use a ControlNet \citep{zhang2023adding} module to condition on our physics signal $\tilde{\pi}$. 
We fine-tune this ControlNet only for the high-noise expert, as this expert is primarily responsible for global structure and low-frequency dynamics \citep{dieleman2024spectral}, which aligns with our physics-planning task.
The ControlNet module clones the first 10 DiT layers from the pretrained Wan2.2, fine-tuning them and feeding their outputs to the frozen base model via zero-convolutions.
We encode the goal force prompt $\pi$ using the frozen Wan2.2 encoder and pass the result through a randomly initialized patch embedding layer before feeding it to the ControlNet DiT layers.
We fine-tune the model for 3,000 steps with an effective batch size of 4 (1 per device on 4 NVIDIA 80GB A100s), which completes in under 48 hours. We use videos of 81 frames at 16 FPS during training and inference.

\section{Experimental Comparisons}
\label{sec:quant_results}

\subsection{Comparison to Text-Only Baselines}

To evaluate Goal Force, we first compare to baselines that use text-only conditioning. 
We create a new benchmark of 75 challenging scenes curated from permissively licensed web sources $\{$Pexels, Pixabay, Unsplash$\}$ as well as generative models $\{$Nano-banana, GPT-Image-1$\}$. 
We then conduct a 2AFC human study ($N=40$) on Prolific, comparing our full Goal Force model against those baselines.

\noindent\textbf{Baselines.} We compare against two models:
\begin{enumerate}
    \item \emph{Text-only (Zero-shot):} Wan2.2 base model, prompted with a text suffix, \eg, \emph{``...a golf ball rolls across the grass, colliding with another ball. The secondary object is moved with very strong force to the left.''}.
    \item \emph{Text-only (Fine-tuned):} Our ControlNet architecture finetuned on our synthetic data, but with the physics control signal zeroed out, relying only on the text suffixes provided during training.
\end{enumerate}

\noindent\textbf{Human Study for Generalization.}
Our benchmark spans four categories of increasing generalization from our training data: (1) two-object collisions (cantaloupes, pendulum striking object, pool ball, rubber duck toys in water, bars of soap, soccer balls, softballs), (2) multi-object collisions (ball colliding with domino, golf balls, tennis balls) (3) human-object interaction (hand interacting with ornaments, toy car; we also include in this category a dog interacting with a ball, and a cat knocking over a chess piece), and (4) tool-object interaction (golf club hitting golf ball, and a fork touching a dome of jello). 
Participants evaluated videos on three axes: \emph{Goal Force Adherence} (Does the video accomplish the specified goal?), \emph{Realistic Motion}, and \emph{Visual Quality}.

Table~\ref{tab:human_study} compares the performance of Goal Force against the text-only baselines.
These results demonstrate that our model outperforms both baselines on goal force adherence, demonstrating that the text prompt is not sufficient, confirming that the explicit physics control signal is critical for solving the task. 
The results also demonstrate that this goal force adherence is achieved with minimal degradation of visual quality and motion realism.
Despite training only on synthetic balls, dominos, and a single flower, our model generalizes effectively, enabling complex, out-of-domain interactions like tool use and human-object planning, as visualized in Figure~\ref{fig:teaser}.

\subsection{Comparison to Prior Methods}

The Goal Force prompting task is new, and prior force-conditioned methods (\eg, PhysGen \citep{liu2024physgen}, PhysDreamer \citep{zhang2024physdreamer}, and Force Prompting \citep{gillman2025forcepromptingvideogeneration}) are not designed to solve it. These models can only simulate a \emph{direct} force (the effect), not plan the \emph{antecedent} action required to achieve a \emph{goal} force (the cause).
As shown qualitatively in Figure~\ref{fig:prior_methods}, when given a goal force prompt, those prior methods misinterpret it as a direct, non-causal poke on the target object. 
Similarly, motion-conditioned models like ToRA \cite{zhang2025tora} can follow a specified trajectory but fail to adhere to causality, often moving the target object before an antecedent event (like a hand) arrives.
A qualitative comparison against these prior works is provided in Figure~\ref{fig:prior_methods}.
Also, while prior methods cannot perform Goal Force prompting, our model is still capable of performing direct Force Prompting (FP).

\section{Goal Force Enables Visual Planning}
\label{sec:visual_planning_probabilistic}

We now evaluate a core claim of our work: that Goal Force enables a form of visual planning. 
We test this by analyzing three key properties of the generated plans: their physical accuracy, their diversity, and their awareness of privileged physics information such as mass.

\subsection{Visual Plans are Accurate}

\begin{figure*}[t]
  \centering
     \vspace{-90pt}
   \includegraphics[width=\linewidth]{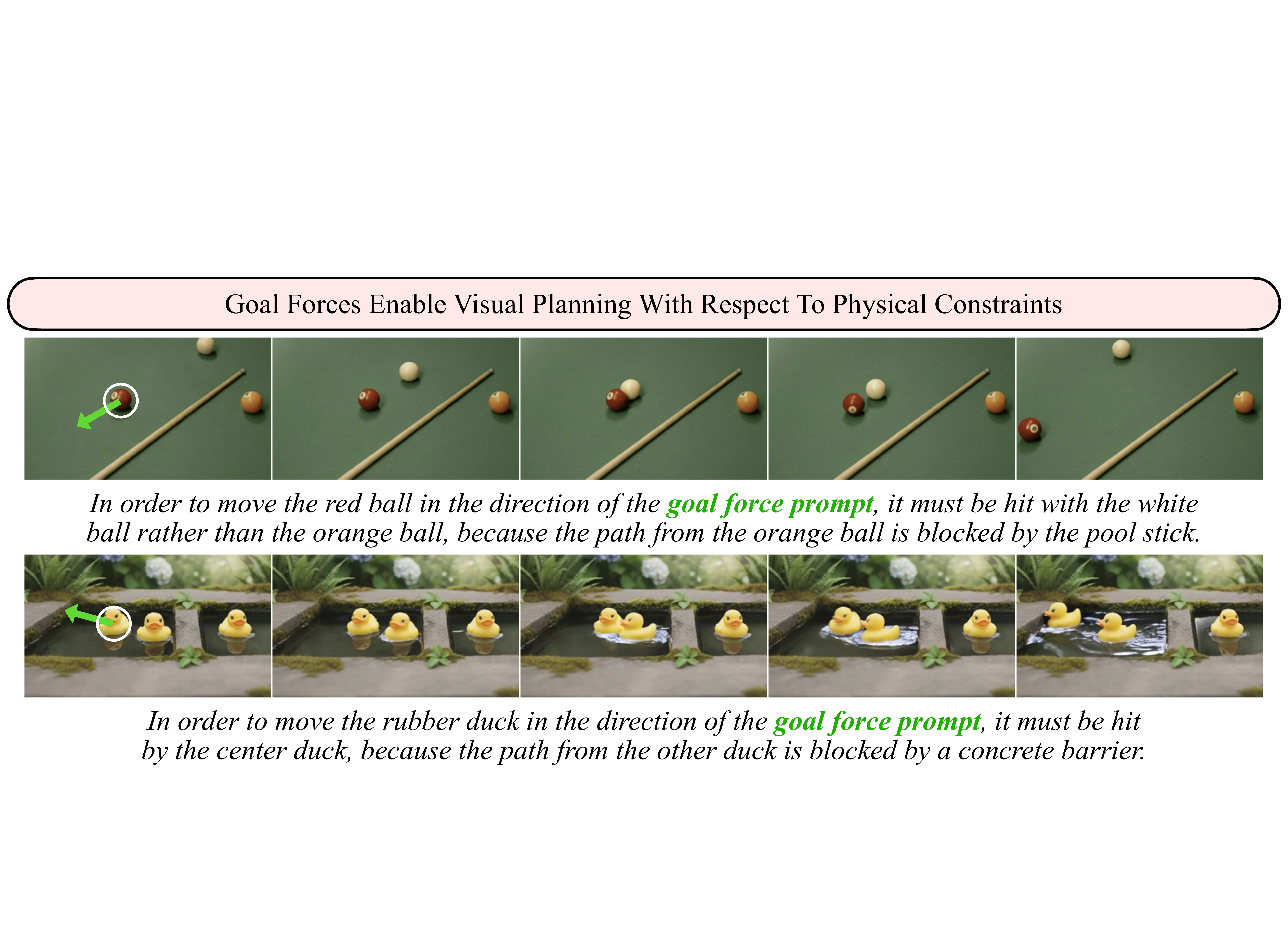}
   \vspace{-80pt}
   \caption{\textbf{Given a goal force prompt, the model chooses the physically correct way to execute it.}
   Top: even though there exist multiple plausible initiators, the model correctly selects the white ball as the initiator to achieve the desired force on the target. 
   Bottom: With multiple plausible rubber ducks that could initiate the force, the model selects the initiator that is not blocked by a physical barrier.\vspace{-1em}}
   \label{fig:probabilistic_visual_plans_accuracy}
\end{figure*}

We first test if the model's visual planning adheres to physical constraints. 
We create a benchmark of 22 scenes containing ``natural blockers'' (Figure~\ref{fig:probabilistic_visual_plans_accuracy}), where each scene features target objects in a contextually appropriate environment, and distractor objects are physically constrained from initiating the goal force. A successful plan requires the model to identify and select a valid, unconstrained object to execute the causal chain.

For each scene, we generate 50 videos. 
To isolate the planning logic from the base video diffusion model's artifacts, we filter out trials exhibiting stochastic visual degradation (\eg, object hallucination) prior to analysis. 
We define \textbf{accuracy} as the percentage of valid trials where the goal force is initiated by the correct, unconstrained object, rather than by a distractor or through spontaneous, non-causal motion.

\begin{table}[t]
\centering
\caption{\textbf{Visual planning accuracy across scenes.} Our model achieves a high success rate in selecting a physically valid force initiator across diverse, complex scenarios. List of all scenes is in Table \ref{tab:vp_scene_accuracy_appendix}.}
\label{tab:vp_scene_accuracy}
\begin{tabular}{@{}lccc@{}}
\toprule
\textbf{Scene}            & \textbf{\# Valid} & \textbf{\# Success} & \textbf{\% Accuracy} \\
\midrule
Pool              & 49                    & 48                     & 97.96                     \\
Paper Balls       & 50                & 49                  & 98.00                 \\
Kitchen Lemon     & 50                & 50                  & 100.00                \\
Coffee Cups       & 44                & 41                  & 93.18                 \\
Duckie            & 40                    & 34                     & 85.00                     \\
Accessories       & 50                & 47                  & 94.00                 \\
Curling Stones    & 49                & 37                  & 75.51                 \\
Rubik\textquotesingle s Cube      & 49                & 46                  & 93.88                 \\
\bottomrule
\end{tabular}
\end{table}

\begin{figure*}[t]
  \centering
  \includegraphics[width=\textwidth]{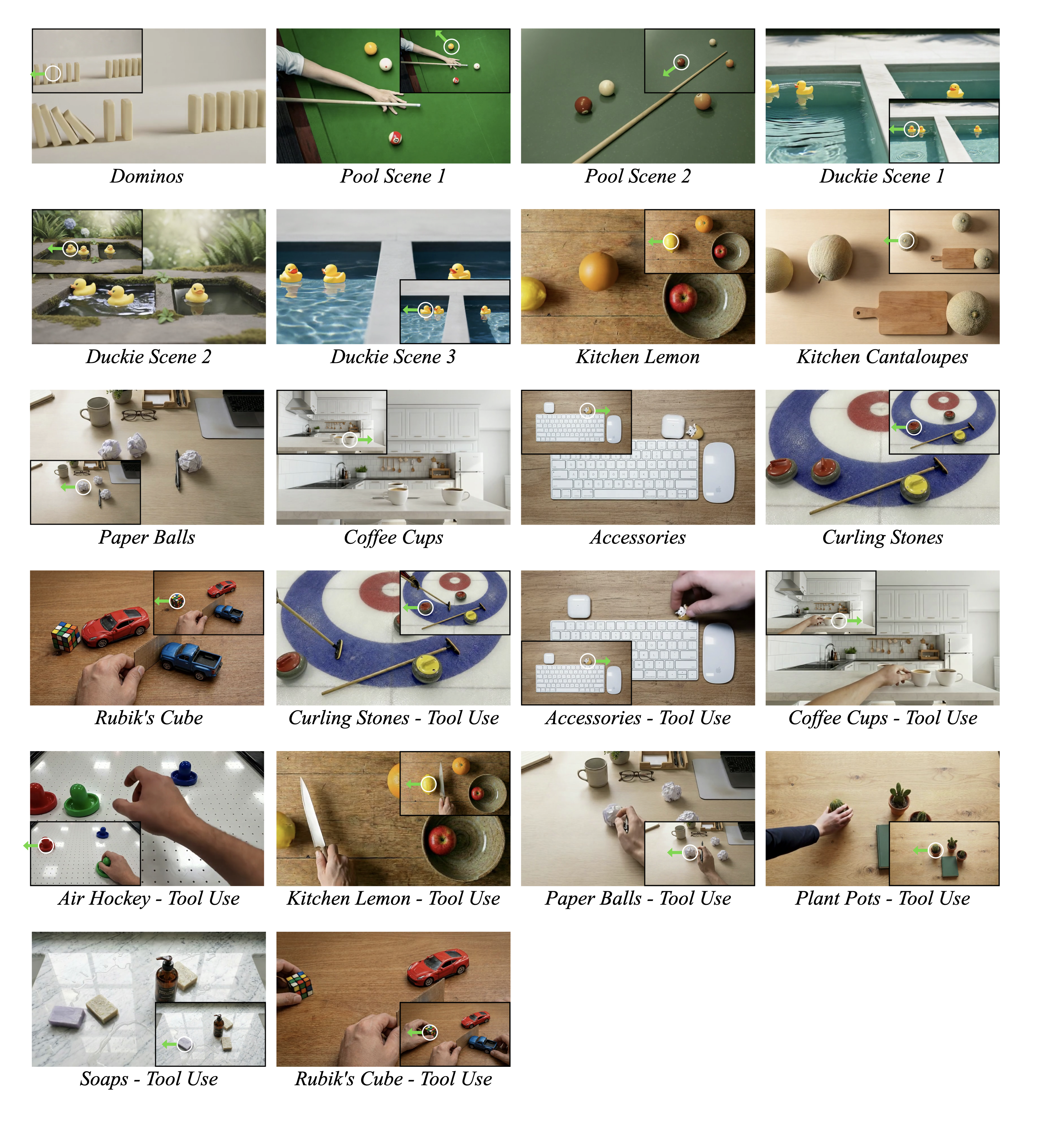}
  \caption{Visualization of all visual planning test scenes. The inset shows the initial state, where the green arrow indicates the goal force, while the larger image shows a valid outcome.}
  \label{fig:vp_scene_visualizations}
\end{figure*}

\noindent\textbf{Results.} We report accuracy for selected scenes in Table~\ref{tab:vp_scene_accuracy}. 
Results for all test scenes are included in Table~\ref{tab:vp_scene_accuracy_appendix}, and Figure~\ref{fig:vp_scene_visualizations} presents visualizations of each scene.  
A random baseline achieves at most 33.3\% accuracy given our distractor design.
The model demonstrates strong physical reasoning across most of the scenes. In the pool example (Figure~\ref{fig:probabilistic_visual_plans_accuracy}, top), a stick blocks the orange ball. Our model correctly selects the white ball as the initiator in 98\% of valid trials. 
On the rubber duckie benchmark (Figure~\ref{fig:probabilistic_visual_plans_accuracy}, bottom), it selects the correct initiator. 
We observe that most failure cases involve the target object moving spontaneously, rather than the model choosing an incorrect, constrained initiator.
We also observe this trend of physically grounded visual planning generalizes to other natural scenarios, including the ones shown in Figure~\ref{fig:teaser}.

\subsection{Visual Plans are Diverse}

Beyond accuracy, we test if our model produces a \emph{diverse} set of valid plans rather than suffering from mode collapse. We design a multi-modal task: a line of six dominos where the goal is to topple the rightmost (sixth) domino block. This goal can be achieved by initiating a chain reaction from any of the five preceding dominos. 
A deterministic model would repeatedly target the same domino, whereas we hypothesize Goal Force will sample from the full distribution of valid plans.

% A non-diverse or deterministic model would exhibit mode collapse, targeting the same domino repeatedly. 
% We hypothesize that our Goal Force model will instead sample from a diverse distribution of valid initial actions.
To quantify this, we propose a diversity metric $\delta(p)$ based on the Jensen-Shannon Divergence (JSD). Let $\hat{p}(x)$ be the empirical probability mass function (PMF) over the set of the $N=5$ targetable dominos, $S = \{0, 1, 2, 3, 4\}$. 
We define our \textbf{diversity metric} as:
\begin{equation}
    \delta(p) = 1 - \mathrm{JSD}(\hat{p} \ \| \ \mathrm{Unif}(S)).
\end{equation}
This metric is normalized to provide an interpretable score. A perfectly diverse model sampling uniformly from all 5 dominos ($\hat{p} = \mathrm{Unif}(S)$) achieves the maximum score of $\delta(p) = 1.0$. Conversely, a fully deterministic model exhibiting complete mode collapse (i.e., $\hat{p}$ is a Dirac delta function on a single domino) yields the baseline score of $\delta(p) \approx 0.39$.

\noindent\textbf{Results.} We present our findings in Table \ref{tab:diversity}. 
Across 26 random seeds, our model achieves a diversity score of 0.6577, significantly higher than the deterministic baseline (0.3900) and distinct from distributions with collapsed support (\eg, $\mathrm{Unif}\{0,1\}$). 
This demonstrates that our model successfully explores a multi-modal distribution of valid plans rather than collapsing to a single solution.

\begin{table}[t]
\centering
\caption{
    \textbf{Diversity metric ($\delta(p)$) scores for the 5-domino task.} 
    Higher is better (Max: 1.0). Our model (0.6577) shows significant 
    diversity compared to the deterministic baseline (0.3900).
}
\label{tab:diversity}
\begin{tabular}{@{}lc@{}}
\toprule
\textbf{Distribution ($p$)}                   & \textbf{Score ($\delta(p)$)} \\
\midrule
\textbf{Our Model (Goal Force)}                      & \textbf{0.6577}              \\
\midrule
\textit{Reference: Unif\{0..4\} (Max diversity)} & 1.0000                       \\
\textit{Reference: Unif\{0..3\}}                 & 0.8920                       \\
\textit{Reference: Unif\{0..2\}}                 & 0.7635                       \\
\textit{Reference: Unif\{0..1\}}                 & 0.6042                       \\
\textit{Reference: Unif\{0\} (Deterministic)}   & 0.3900                       \\
\bottomrule
\end{tabular}
\end{table}

\begin{figure}[t]
  \centering
  \includegraphics[width=\linewidth]{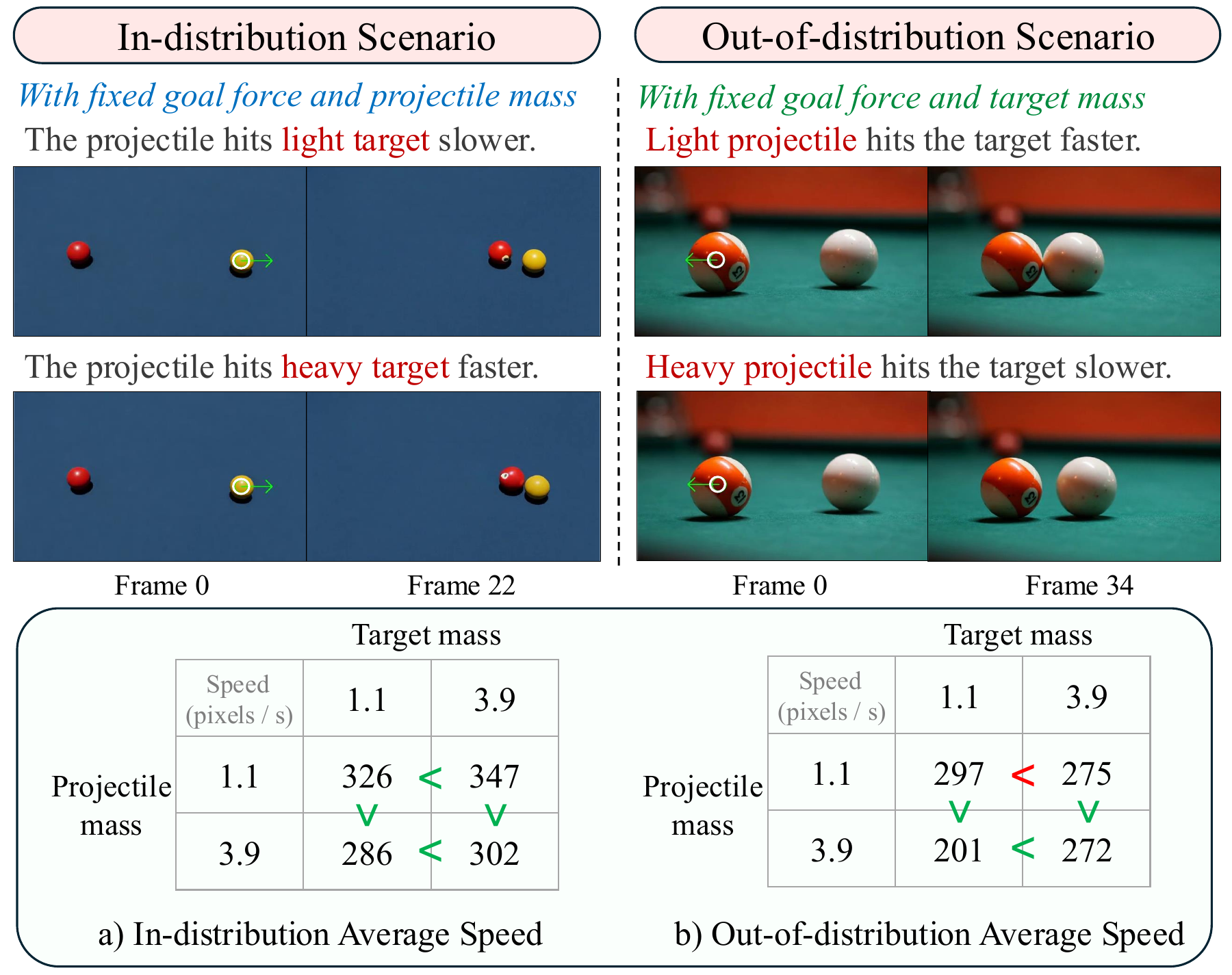}

   \caption{\textbf{Visual plans take advantage of mass information.} We test goal force prompting on in-distribution (left) and out-of-distribution (right) scenarios. In both scenarios, our model can adjust the moving speed of the projectile accordingly when the object masses are changed to cause the desired force magnitude. The direction of the ``$<$'' sign indicates the desired numerical relationship; \textcolor{Green}{green} indicates satisfaction, \textcolor{red}{red} indicates violation.\vspace{-2em}}
   \label{fig:probabilistic_visual_plans_physics}
\end{figure}

\subsection{Visual Plans Leverage Privileged Physics}

Next, we test if the model's visual plans can use privileged mass information provided in the control signal to help guide their plans. 
Our experiments focus on ball collision.
In this setting, a physically-grounded plan must account for mass; for example, achieving a specific goal force on a heavier target requires a stronger impact.

We design a ball collision task with a fixed goal force magnitude, varying the projectile and target masses. 
In Figure~\ref{fig:probabilistic_visual_plans_physics}, we test our model on two such scenarios. One is in-distribution with a scene and viewpoint similar to our training data.
The other features an out-of-distribution background, viewpoint, lighting, and ball size.
We expect the model to learn two principles: (1) if projectile mass is constant, a heavier target requires a faster projectile; (2) if target mass is constant, a heavier projectile can move slower.

To quantitatively measure the ball collision, we use Faster R-CNN~\cite{ren2016fasterrcnnrealtimeobject} to detect the positions of the two balls.
Then we determine the collision time and compute the projectile's moving speed accordingly.
We generate 15 videos for each combination of masses and average the speed over the samples.
We observe that in the in-distribution scenario, the projectile's speeds satisfy all four desired speed magnitude relationships.
In the out-of-distribution scenario, our results satisfy three of them, while the fourth is very close.
This demonstrates the model's capability in leveraging privileged physics information for visual planning.

\section{Conclusion}
\label{sec:conclusion}

We introduce Goal Force, a paradigm that shifts generative video control from specifying a direct force (the cause) to declaring a desired goal force (the effect). 
We demonstrate that by training on simple, synthetic causal primitives, a video model can learn to function as an implicit neural physics planner. 
This enables the model to reason backward from a user-defined goal and generate a physically plausible, antecedent causal chain to achieve it. 
Our key finding is that this planning capability generalizes to complex, out-of-domain scenarios involving tool use and human-object interactions. 
This work represents a step toward interactive world models that can not only simulate a physical reaction but also reason about and plan the actions required to achieve a desired physical outcome.

\vspace{.2em}\noindent\textbf{Acknowledgements:} 
We would like to thank Bill Freeman, Calvin Luo, David Fleet, Miki Rubinstein, and Singh Saluja for useful discussions.
This material is based upon work partially supported by the U.S. National Science Foundation under Cooperative Agreement No. 2433429.
Our research was conducted using computational resources at the Center for Computation
and Visualization at Brown University.

{
    \small
    \bibliographystyle{ieeenat_fullname}
    \bibliography{main}
}

\clearpage
\setcounter{page}{1}
\maketitlesupplementary

\begin{table*}[t]

\begin{center}
\small % Smaller font size than \small
\setlength{\tabcolsep}{3pt} % Reduce column spacing (default is 6pt)

\begin{tabular}{l || c|c|c|c||c|c|c} 
 \textbf{Visual Quality} & 
 Dominos & Pool balls & Stone tower & Wall toy & Orange Rose & White Rose & Tulip\\ 
 \hline\hline
Force Prompting & 90.0\% & 80.0\% & 60.0\% & 50.0\% & 100.0\% & 80.0\% & 60.0\%\\ 
PhysGen & 60.0\% & 100.0\% & 100.0\% & 80.0\% & -- & -- & --\\ 
PhysDreamer & -- & -- & -- & -- & 50.0\% & 50.0\% & 50.0\%\\ 
 \hline\hline
\end{tabular}

\end{center}

\begin{center}
\small % Smaller font size than \small
\setlength{\tabcolsep}{3pt} % Reduce column spacing (default is 6pt)

\begin{tabular}{l || c|c|c|c||c|c|c} 
 \textbf{Force Adherence} & 
 Dominos & Pool balls & Stone tower & Wall toy & Orange Rose & White Rose & Tulip\\  
 \hline\hline
Force Prompting  & 90.0\% & 90.0\% & 80.0\% & 90.0\% & 50.0\% & 60.0\% & 50.0\%\\ 
PhysGen  & 90.0\% & 80.0\% & 80.0\% & 40.0\% & -- & -- & --\\ 
PhysDreamer & -- & -- & -- & -- & 40.0\% & 30.0\% & 60.0\%\\ 
 \hline\hline
\end{tabular}

\caption{\textbf{Human study comparing the Direct Force capability of the Goal Force method to prior works.} 
Numbers indicate the percentage of human pairwise preferences for Goal Force Prompting's direct force capability (i.e. encoding the force in the first channel) over each baseline on each benchmark dataset. 
The results demonstrate that Goal Force achieves consistently higher visual quality, as well as superior force adherence against the majority of baselines. Notably, our method achieves these results without relying on physics simulators or 3D assets at inference, unlike PhysDreamer and PhysGen.
We note that PhysGen models rigid body mechanics, whereas PhysDreamer models oscillations, so they can't be directly compared to one another.}
\label{tab:direct_force_comparison}

\end{center}

\end{table*}

\begin{table*}[t]
\centering
\begin{tabular}{@{}lccc@{}}
\toprule
\textbf{Scene}                     & \textbf{\# Valid} & \textbf{\# Success} & \textbf{\% Accuracy} \\
\midrule
Dominos           & 50                    & 50                     & 100.00                    \\
Pool Scene 1              & 49                    & 48                     & 97.96                     \\
Pool Scene 2              & 22                    & 12                     & 54.55                     \\
Duckie Scene 1           & 40                    & 34                     & 85.00                     \\
Duckie Scene 2            & 37                    & 24                     & 64.86                     \\
Duckie Scene 3            & 41                    & 38                     & 92.68                     \\
Kitchen Lemon     & 50                & 50                  & 100.00                \\
Kitchen Cantaloupes        & 50                & 39                  & 78.00                 \\
Paper Balls       & 50                & 49                  & 98.00                 \\
Coffee Cups       & 44                & 41                  & 93.18                 \\
Accessories       & 50                & 47                  & 94.00                 \\
Curling Stones    & 49                & 37                  & 75.51                 \\
Rubik\textquotesingle s Cube      & 49                & 46                  & 93.88                 \\
Curling Stones - Tool Use& 47                & 45                  & 95.74                 \\
Accessories - Tool Use    & 47                & 40                  & 85.11                 \\
Coffee Cups - Tool Use    & 44                & 44                  & 100.00                \\
Air Hockey - Tool Use              & 45                & 38                  & 84.44                 \\
Kitchen Lemon - Tool Use  & 50                & 44                  & 88.00                 \\
Paper Balls - Tool Use    & 50                & 47                  & 94.00                 \\
Plant Pots - Tool Use    & 46                & 38                  & 82.61                 \\
Soaps - Tool Use          & 48                & 42                  & 87.50                 \\
Rubik\textquotesingle s Cube - Tool Use       & 48                & 45                  & 93.75                 \\
\bottomrule
\end{tabular}
\caption{\textbf{Visual planning results for all test scenes.} We also report the visual planning accuracy for diverse tool use scenarios, where success is defined as correctly using tools to achieve the specified goal force in the presence of distractors. The results show that the model achieves a high success rate in most cases across diverse and complex scenarios.}
\label{tab:vp_scene_accuracy_appendix}
\end{table*}

\section{Additional Experiment Details}
\label{app:additional_experiment_details}

\subsection{Comparison to Prior Works: Direct Force Prompting Quantitative Comparison}
\label{app:direct_force_comparison}

We encode the goal force prompt in the second channel of the control signal, and we encode the direct force prompt (which is a similar task to PhysDreamer \citep{zhang2024physdreamer}, Force Prompting \citep{gillman2025forcepromptingvideogeneration}, and PhysGen \citep{liu2024physgen}) in the first channel of the control signal.
In Table~\ref{tab:direct_force_comparison} we compare the ``Direct Force Prompting'' capability of our model to those three models via a 2AFC human study ($N=10)$ conducted on Prolific.
We gathered two benchmarks: 
    a PhysGen benchmark, consisting of four scenes highlighted on that work's project page;
    as well as a PhysDreamer benchmark, consisting of three scenes highlighted on that work's project page.
We compare our model to PhysGen and Force Prompting on the PhysGen benchmark, and we compare our model to PhysDreamer and Force Prompting on the PhysDreamer benchmark.
Note that PhysGen models rigid body mechanics, whereas PhysDreamer models oscillations.

\subsection{Synthetic Data Generation}
\label{subsec:synthetic-extended}

In this section, we provide an in-depth discussion of the methods and specific parameters utilized for generating our synthetic training data. 
This data is used to train the Goal Force model to act as an implicit neural physics planner.

For all synthetic datasets, a key step in creating the multi-channel control signal $\tilde{\pi}$ is the projection of 3D forces and object properties onto the 2D image plane. We use the camera's parameters to map 3D force vectors and object world coordinates into 2D pixel coordinates, enabling us to accurately model physical interactions within the video frames.

\subsubsection{Dominos Dataset}
We generated 3k videos of domino chain reactions using Blender. The setup models a causal chain where an initial direct force on one domino results in a predictable goal force on a downstream target domino.

To ensure diversity and robustness, we randomized the following parameters per video:
\begin{itemize}
    \item \textbf{Domino Count:} Uniformly sampled from $\mathrm{Unif}\{3, \dots, 10\}$.
    \item \textbf{Scene Geometry:} Randomized placement and orientation of the domino line.
    \item \textbf{Causality:} Choice of the initial target domino and the direction of the chain reaction (i.e., hitting the domino in front or behind).
    \item \textbf{Visuals:} Randomized camera position, ground textures (from $42$ Polyhaven options), and High Dynamic Range Images (HDRIs) for lighting and background (from $50$ Polyhaven options).
    \item \textbf{Force Magnitude:} Continuous values from $[0, 1]$, where $0$ represents the minimum force required for the domino to topple and $1$ represents a maximal, strong impulse.
\end{itemize}
Each video is accompanied by a JSON file that records the names of the initial and adjacent contact dominos, along with the complete 2D pixel coordinates for all dominos across every frame.

\subsubsection{Rolling Balls Dataset}
This dataset comprises 6k videos generated in Blender, split into two primary categories to capture both collision and non-collision causal interactions:
\begin{enumerate}
    \item Collision Set ($4.5$k videos): A ``projectile'' ball, acted upon by an unseen point force, is aimed to collide with one specific ``target'' ball within a group of initially stationary ``distractor'' balls.
    \item Non-Collision Set ($1.5$k videos): The projectile ball is aimed such that it misses the target ball.
\end{enumerate}

For the Collision Set, we ensured a diverse range of physical scenarios by randomizing:
\begin{itemize}
    \item \textbf{Ball Count:} $\mathrm{Unif}\{3, \dots, 9\}$.
    \item \textbf{Physical Properties:} Ball colors, ball masses $\mathrm{Unif}(1.0, 4.0)$ kg, and all ball positions.
    \item \textbf{Visuals:} Randomized camera position and ground textures.
    \item \textbf{Force Calculation:} To guarantee collision, a minimum required force is calculated based on the projectile mass, distance to the target, and a randomized collision time ($\mathrm{Unif}(2.5, 4.5)$ seconds). This minimum force is scaled by $\mathrm{Unif}(1.2,1.6)$ to introduce physical variation.
\end{itemize}

The collision videos are evenly split between straight-on and indirect collisions.
For both types, the script first calculates the precise angular window necessary for the projectile to hit the target.
\begin{itemize}
    \item For straight-on collisions, the force is aimed directly at the center of this calculated angular window.
    \item For indirect collisions, the force angle is randomly sampled within this window, resulting in an off-center strike. 
\end{itemize}
This mixed-collision approach helps the model learn diverse post-collision behaviors.

For the Non-Collision Set, we randomized: ball quantity ($\mathrm{Unif}\{3, \dots, 5\}$), ball textures, positions, camera angle, ground textures, target ball selection, force angle ($[0, 360^\circ)$), and force magnitude ($[0, 1]$).

For all ball videos, a JSON file records initial 2D/3D coordinates and physics parameters. For the videos featuring indirect collisions, we also save the complete 2D pixel trajectory of the target ball. For the non-collision videos, we save the final 2D trajectory angle of the projectile ball.

\subsubsection{Plants Dataset}
This dataset, generated using PhysDreamer \cite{zhang2024physdreamer} (which integrates 3D Gaussians and a physics simulator), focuses on non-rigid body dynamics. The videos show a plant (carnation) swaying after being subjected to a direct force. We randomized the following parameters:
\begin{itemize}
    \item \textbf{Initial Conditions:} Camera position and initial object configuration.
    \item \textbf{Force Application:} Contact points, force angles, and force magnitudes in $[0, 1]$, where $0$ is a gentle poke and $1$ is a strong impulse.
\end{itemize}

\subsection{Ablation studies}\label{app:ablation_studies}

\subsubsection{How important is the mass channel?}

We find that masking the mass channel during training and relying on text instead for mass leads to worse performance than reported in Figure~\ref{fig:probabilistic_visual_plans_physics}.

\subsubsection{How important is the direct force channel?}

We find that masking the direct force channel during training causes the model to fail on some of the complex, out-of-domain causal chains, such as human-object interactions.

\subsubsection{How specific does the text prompt need to be?}

% \ZWmc, \Sevenjqn: \setuldepth{a}\ul{\emph{Method is biased by text in terms of causality.}} 
We conducted ablation studies to confirm that the action itself does not need to be specified in the text prompt (\eg, ``the dog paw \emph{causes a ball to move}'' works just as well as ``the dog paw \emph{nudges the ball}''). 
When the source of action is unspecified, the model resorts to its pre-trained prior to choose a cause, which may sometimes be an ``invisible'' force.
We highlight that, even when the source of action is specified in the text prompt, the low-level physical dynamics are not (\eg, we never specify to ``hit the ball at that angle with that force'').
Additionally, the ability to specify the interaction source and type is a desirable feature (\eg, for robotics applications), as it allows users to provide the embodiment and the high-level plan information.

\end{document}